\documentclass[10pt,twocolumn,letterpaper]{article}

\usepackage{iccv}
\usepackage{times}
\usepackage{epsfig}
\usepackage{graphicx}
\usepackage{amsmath}
\usepackage{amssymb}

\usepackage{times}
\usepackage{epsfig}
\usepackage{graphicx}
\usepackage{amsmath}
\usepackage{amssymb}

\usepackage{booktabs}
\usepackage{comment}
\usepackage{amsmath,amssymb} 
\usepackage{color}
\usepackage{epsfig}
\usepackage{subfigure}
\usepackage{booktabs}
\usepackage{multirow}
\usepackage{array}
\usepackage{setspace}
\usepackage{mathrsfs}
\usepackage{algorithm}
\usepackage{algorithmic}
\usepackage{caption}
\usepackage{color}

\newcommand{\tabincell}[2]{\begin{tabular}{@{}#1@{}}#2\end{tabular}}

\usepackage[pagebackref=true,breaklinks=true,letterpaper=true,colorlinks,bookmarks=false]{hyperref}

 \iccvfinalcopy 

\ificcvfinal\fi

\begin{document}

\title{OCM3D: Object-Centric Monocular 3D Object Detection}

\author{
Liang Peng$^{1, 2}$ 
\quad Fei Liu$^{2}$ 
\quad Senbo Yan${^{1,2}}$ 
\quad Xiaofei He${^{1,2}}$ 
\quad Deng Cai $^1$
\quad \\
\textsuperscript{\rm 1}State Key Lab of CAD\&CG, Zhejiang University \\
\textsuperscript{\rm 2}FABU Inc. \\
{\tt\small  \{pengliang, senboyan\}@zju.edu.cn \quad  \{xiaofei\_h,dengcai78\}@qq.com \quad \{liufei\}@fabu.ai}
}

\maketitle

\ificcvfinal\thispagestyle{empty}\fi

\begin{abstract}
	Image-only and pseudo-LiDAR representations are commonly used for monocular 3D object detection. 
	However, methods based on them have shortcomings of either not well capturing the spatial relationships in neighbored image pixels or being hard to handle the noisy nature of the monocular pseudo-LiDAR point cloud. 
	To overcome these issues, in this paper we propose a novel object-centric voxel representation tailored for monocular 3D object detection. 
	Specifically, voxels are built on each object proposal, and their sizes are adaptively determined by the 3D spatial distribution of the points, allowing the noisy point cloud to be organized effectively within a voxel grid. 
	This representation is proved to be able to locate the object in 3D space accurately. 
	Furthermore, prior works would like to estimate the orientation via deep features extracted from an entire image or a noisy point cloud. 
	By contrast, we argue that the local RoI information from the object image patch alone with a proper resizing scheme is a better input as it provides complete semantic clues meanwhile excludes irrelevant interferences. 
	Besides, we decompose the confidence mechanism in monocular 3D object detection by considering the relationship between 3D objects and the associated 2D boxes. 
	Evaluated on KITTI, our method outperforms state-of-the-art methods by a large margin. The code will be made publicly available soon.
\end{abstract}

\section{Introduction}

	3D object detection is of great concern in autonomous driving, and a variety of methods have been proposed. 
	These methods can be broadly divided into two categories: LiDAR-based methods \cite{VoxelNet,PP,PVRCNN} and camera-based methods \cite{Deep3DBBox,OFTNet,FQNet, StereoRCNN,DSGN}. 
	LiDAR-based methods usually offer satisfactory performances as the LiDAR point cloud can provide accurate depth measurement of the scene. 
	However, the limited working range and the high price of LiDAR devices are the main disadvantages of this category of methods. 
	As an alternative, camera-based methods, although still far from satisfactory in terms of the detection performance, rely only on camera sensors that are much cheaper and maturer. 
	Therefore, these approaches, especially the monocular ones \cite{Mono3D,ROI10D,MonoGRNet,M3D,RTM3D}, have drawn increasing attention from both industry and academia.

	Previous image-only based monocular methods lack explicit knowledge about the depth dimension, having difficulties in predicting objects in 3D space precisely.
	To this end, many recent state-of-the-art monocular 3D object detection methods \cite{D4LCN,PseudoLidar, Mono_PseudoLidar,AM3D} utilize the estimated depth map to make use of depth information.  
	Some of them \cite{PseudoLidar, Mono_PseudoLidar} convert the depth map to a pseudo-LiDAR point cloud and then perform detection on it. 
	However, most pseudo-LiDAR based methods use existing 3D detectors designed purposely for the accurate LiDAR point cloud, failing to capture the character of the highly noisy monocular pseudo-LiDAR with the long-tail defect and thus resulting in suboptimal performance.

	To address the above problems, we propose a novel object-centric voxel representation.
	For every 2D object proposal generated by the 2D detector, we use the estimated depth map to convert pixels within the 2D box into a point cloud and build adaptive voxels on it, whose sizes depend on the 3D spatial distribution of points.
	With this voxel building scheme, the region with dense points that belongs more likely to an object will be partitioned with dense voxels, making more information to be encoded. 
	On the contrary, the region with sparse points that are more likely outliers will be partitioned with sparse voxels, enabling more irrelevant interferences to be eliminated.
	Compared to pseudo-LiDAR, this novel voxel representation is tailor-made for monocular imagery and against the highly noisy point cloud, encoding information from the transformed point cloud and the RGB image effectively.

	Another contribution is the investigation concerning the input for orientation estimation. 
	Orientation plays a critical role in 3D object detection, tracking, and trajectory prediction. 
	Prior works usually perform orientation estimation on deep features extracted from an entire image \cite{OFTNet,AP40,M3D,D4LCN} or a point cloud \cite{PseudoLidar, Mono_PseudoLidar,AM3D}.
	However, the orientation only hinges on the object's appearance on the local image.
	Semantics from the region outside the object is unnecessary, which may interfere or even overwhelm the local orientation features that matter. 
	For this reason, we argue that the object image patch alone cropped by a 2D box is preferable as input for orientation prediction.
	We are the first one that points out this.
	Also, as detailed in Sec. \ref{sec:orient}, 	the cropped image patch needs to be properly processed to eliminate orientation ambiguity. 
	Experiments show that our method performs best, and even comparably to the method \cite{FPointnet}  relying on LiDAR input on the pedestrian category.

	Besides, an object with high 2D detection confidence may be hard to be located accurately in 3D space, \eg, occluded, truncated, or distant objects, suggesting that the 2D confidence should not be naively employed in 3D detection.
	However, for 3D detection methods based on a 2D detector, it is hard to learn 3D confidence without explicit labels during training. 
	Consequently, most SOTA methods \cite{PseudoLidar,Mono_PseudoLidar,AM3D}  directly apply the 2D confidence as the 3D detection confidence score. 
	To resolve this problem, we propose to decompose the 3D confidence mechanism into the confidence in 2D detection and the lifting hardness of an object from 2D to 3D, where the former can be easily obtained from a 2D detector.
	 The lifting hardness is measured by the relationship between the 2D box and 3D box projections as we believe that objects are worthy of high confidences when the 2D and 3D detections fit tightly on the image plane.
	Note that the proposed method can be plunged into any 2D-detector-based method without training and significantly boost the performance.

	Finally, we discover that the training set in Eigen split \cite{Eigen} that is widely adopted for training depth estimation networks overlaps the KITTI 3D object detection validation set. 
	This data leakage results in that almost all the depth-based monocular 3D object detection methods are overrated in the KITTI validation set.
	To remedy this problem, we introduce a new dataset split. 
	In conclusion, our main contributions can be summarized as follows:
	\begin{itemize}
	\item We propose a novel object-centric voxel representation, which effectively encodes the noisy point cloud and the RGB image. 
	\item We point out that a cropped object image patch is preferable as input compared to an entire image for orientation prediction.
	\item A novel decomposed 3D confidence mechanism is designed for monocular 3D object detection, taking both 2D confidence and the lifting hardness from 2D to 3D into consideration.
	\item A widely existing data leakage issue is pointed out and remedied.
		 Also, extensive experiments show that we achieve state-of-the-art performance on KITTI monocular 3D detection benchmark with a significant margin.
	\end{itemize}

	\begin{figure*}[t]
		\begin{center}
			\begin{minipage}[t]{1\textwidth}
				\centering
				\includegraphics[width=1\textwidth]{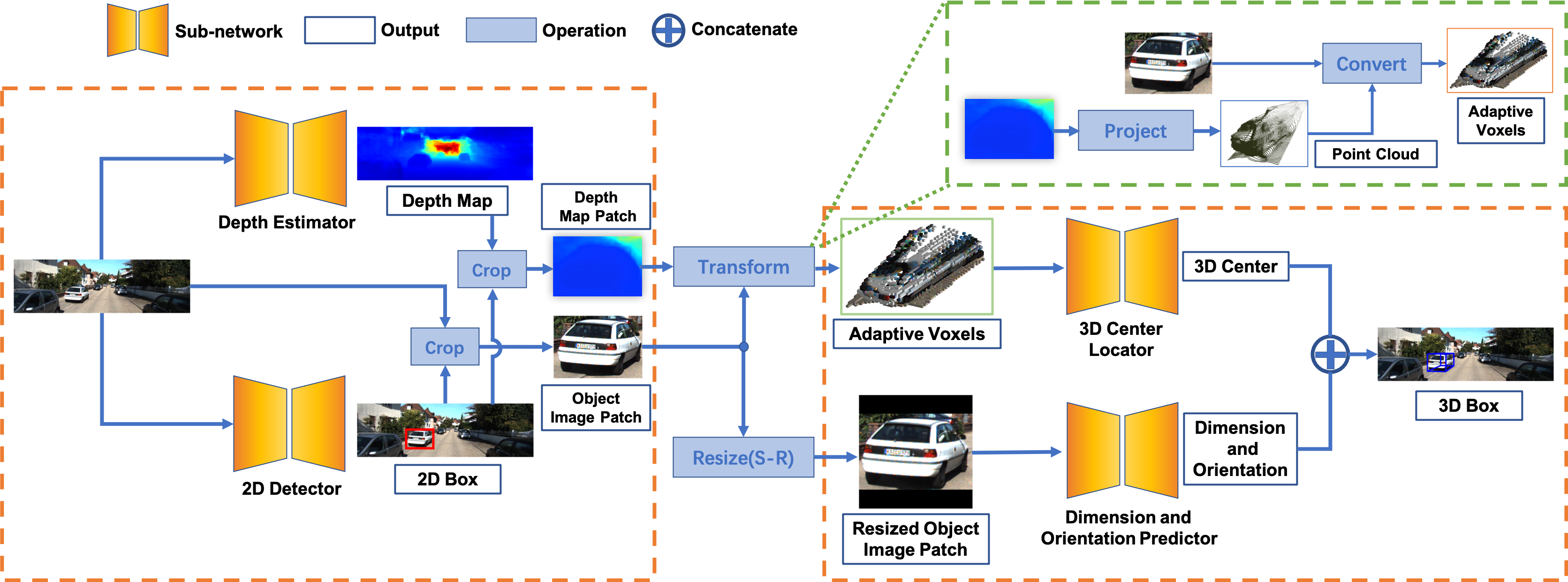}
			\end{minipage}
		\end{center}
		\caption{OCM3D framework. OCM3D consists of four subnetworks for 2D detection, depth estimation, 3D center location, and dimension and orientation prediction, respectively. The 2D boxes and depth map are firstly produced by off-the-shelf networks. For each object, the estimated depth map and 2D box are used to construct adaptive voxels  (Sec. \ref{subsec:adp}), on which a 3D center locator predicts the object's 3D center. By utilizing another subnetwork, the object's dimension and orientation are predicted from the cropped image patch (Sec. \ref{sec:orient}). Best viewed in color with zoom in.}
		\label{fig:full_network}
	\end{figure*}

\section{Related Work}
\noindent
{\bf LiDAR-based Methods:} 
	Most state-of-the-art 3D object detection methods \cite{PART,PP,SA-SSD,3D-CVF,Point-gnn,3dssd,HVNet,Epnet} employ LiDAR as it can provide accurate point clouds, in which voxel-based methods are utmost related to our method. 
	VoxelNet \cite{VoxelNet} divides the LiDAR point cloud into a voxel grid with a fixed voxel size, a group of points that belong to the same voxel is fed into a fully connected network to form the unified feature representation. 
	The 2D convolution of these higher-order voxel features is carried out to obtain detection results.
	Following voxel-based approaches \cite{VoxelFPN,PVCNN} follow this line of thought. 
	Such designs tailored for accurate LiDAR points are not suitable for monocular methods since the point cloud transformed from the predicted depth map is much more inaccurate. 
	Points of the same object distribute chaotically within the whole 3D space (\eg, 3D points transformed from neighboring pixels can appear far away in 3D space, even they are supposed to be much close). 
	CNNs are hard to capture local clues from voxels built with the way above.
	
 ~\\
 \noindent
{\bf Representations for Monocular Methods:}
	Monocular methods can be roughly divided into image-only based methods and depth map based methods according to representations.
	Previous image-only based methods usually take prior knowledge as guidance in training or as constraints in post-processing.
	For example, Mono3D \cite{Mono3D} uses semantic and context priors as guidance to generate 3D proposals, require extra semantic and instance segmentation networks. 
	M3D-RPN \cite{M3D} uses different convolution kernels in row-spaces, trying to explore different features in different depth ranges and RTM3D \cite{RTM3D} predicts perspective key points to refine initial guess by solving a constrained optimization problem. 
	These image-only based methods have difficulties in exploiting the hidden depth information, remaining room for improvement. 
	To make use of depth information explicitly, many methods use the depth map produced by a depth estimating network.
	D4LCN \cite{D4LCN} uses different kernels generated by depth maps, but not full use of the explicitly spatial relationship. 
	Pseudo-LiDAR \cite{PseudoLidar, Mono_PseudoLidar} converts image-based depth maps to point clouds to mimic the LiDAR signal, and then directly using LiDAR-based 3D detectors.
	On this basis,  AM3D \cite{AM3D} augments this method by embedding RGB values to generate attention maps for multi-modal features fusion. 
	However, pseudo-LiDAR based methods ignore the significant gap between the transformed noisy point cloud and the accurate LiDAR point cloud.
	Furthermore, PatchNet \cite{PatchNet} points out that the effectiveness of pseudo-LiDAR comes from the coordinates transform.  
	It organizes the pseudo-LiDAR as the image representation and utilizes existing 2D CNN designs to extract deep features. 
	However, it does not consider utilizing the 3D spatial relationship inside the object.

 \noindent
{\bf Orientation Estimation:}
	Many approaches use the entire image directly for orientation estimation, such as \cite{OFTNet,AP40,M3D,D4LCN}. 
	The global semantics extracted from the region outside the object may interfere or even overwhelm the local semantic that matters for orientation estimation. 
	Pseudo-LiDAR \cite{PseudoLidar, Mono_PseudoLidar} performs prediction on the point cloud, which is hard to infer orientations of objects with noisy points. 
	Deep3DBBox \cite{Deep3DBBox} takes cropped object image patches as input, using MultiBin architecture for orientation regression. 
	However, they process the image patch by using naive resizing operation, making the orientation ambiguous. 
	We also utilize the object image patch and introduce another simple yet effective resizing scheme to replace the naive resizing without damaging any semantic clues. 
	Based on this proper resizing scheme, we argue that the object image patch alone provides adequate features to predict orientations.

 \noindent
{\bf Monocular 3D Object Detection Confidence:}
	Previous works in monocular 3D object detection realize the need for 3D box confidence.  
	FQNet \cite{FQNet} employs 3D IoU loss calculated from the predicted 3D box and the ground truth, mainly considering the 3D object dimensions. 
	MonoDIS \cite{AP40} takes advantage of the 3D box loss, represents the confidence of 3D detection. 
	As a video-based method, the recent concurrent work, Kinematic 3D \cite{Kinematic3D} , introduces a self-balancing confidence loss, re-balancing hard 3D boxes and focusing on relatively achievable samples. 
	All of these methods consider the confidence towards loss function. 
	By contrast, our method takes the physical projections into consideration, gains promising performances, and can be adopted conveniently  by other methods.

\section{Methods}
\subsection{Overview}
	Our object-centric monocular 3D detection framework (OCM3D) can be divided into two stages.
	As shown in Fig. \ref{fig:full_network}, stage one contains 2D detection and depth map estimation, producing 2D boxes and the image-based depth map by two off-the-shelf networks \cite{FPointnet,BTS}, as a common practice \cite{PseudoLidar, Mono_PseudoLidar,AM3D,PatchNet}.
	Our method mainly focuses on stage two that comprises object 3D center location and dimension \& orientation prediction.

	 In stage two, dimensions and orientations are predicted via cropped object image patches with a proper resizing scheme (Sec. \ref{sec:orient}). 
	At the same time, with the addition of the depth map, adaptive voxels are constructed (Sec. \ref{subsec:adp}) and then fed into a 3D fully convolutional network \cite{FCN,UNet,3D_UNet} (see the supplementary material). 
	The network outputs a 3D heat map to obtain the 3D center, since we formulate the 3D center location problem as a key-point prediction task (Sec. \ref{subsec:adp}). 
	Finally, the 3D detection confidence is determined by the 2D detection confidence and the lifting hardness from 2D to 3D (Sec. \ref{sec:conf}).

\subsection{Adaptive Voxels}{\label{subsec:adp}}

\begin{figure}
\begin{minipage}[b]{0.48\textwidth} 
\centering 
			\subfigure{
				\label{fig_1}
				\centering
				\includegraphics[width=1\linewidth]{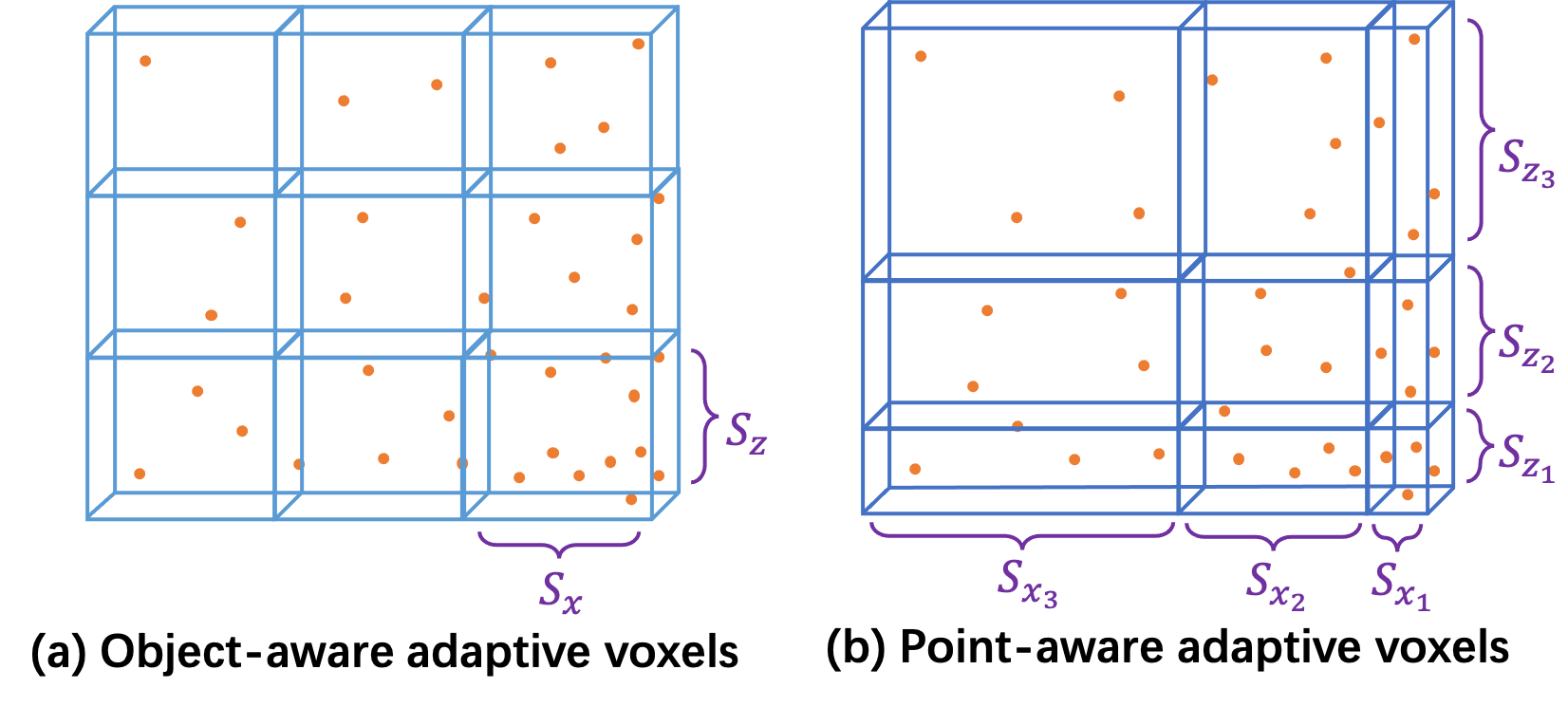}      
			}
		\caption{Illustration of different adaptive voxels. Only $X$ and $Z$ axis are shown for convenience. Orange dots represent the RoI point cloud, $S_x, S_z$ refer to the voxel sizes along $X$ and $Z$ axis, and each small cube is a voxel.}
		\label{fig:interval}
\end{minipage}
\end{figure}

\begin{figure}
\begin{minipage}[b]{0.48\textwidth} 
\centering 
			\subfigure{
				\centering
				\includegraphics[width=1\linewidth]{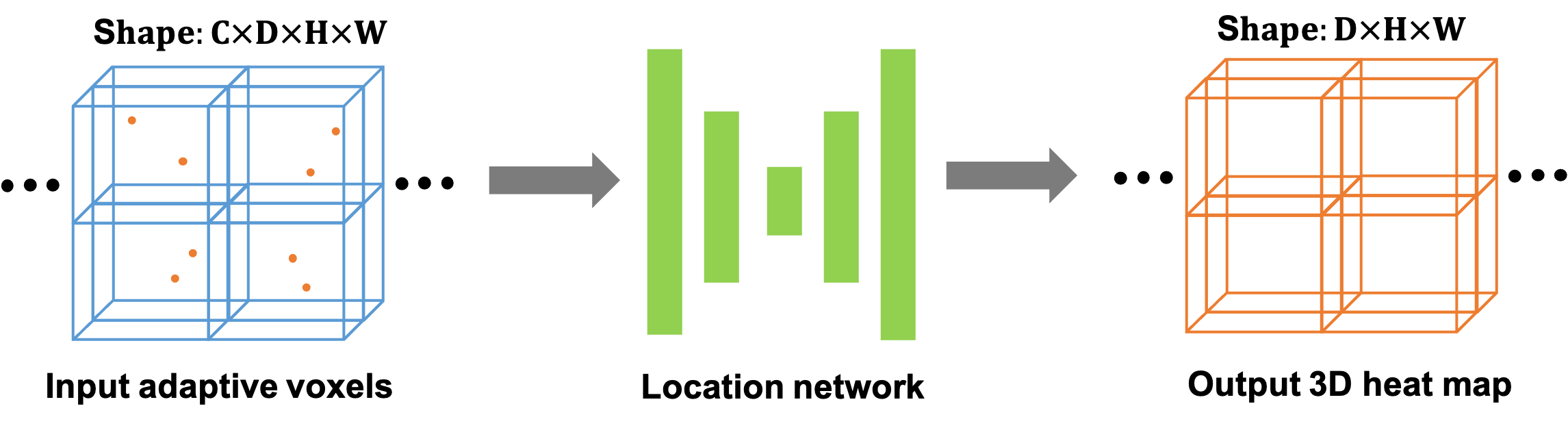}      
			}
		\caption{Illustration of obtaining the 3D heatmap.}
		\label{fig:net2}
\end{minipage}
\end{figure}

	In this section, we describe how to construct adaptive voxels.
	Unlike prior works that divide the entire 3D space into a voxel grid with the fixed voxel size, we first partition the 3D space to obtain an RoI 3D space, which contains only the specified object, enabling fine-grained voxels to be constructed for encoding detailed features.
	Thanks to the fine-grained design, RGB values on the image can be utilized directly as initial features for corresponding voxels via the projection,  allowing the network to utilize the object's visual clues and spatial relationship adequately.	
	Therefore the remaining problem is how to customize the voxel size for every object to fit their noisy and varied point cloud distribution.
	
	Specifically, as shown in Fig. \ref{fig:full_network} and Fig. \ref{fig:interval}, given a 2D object proposal in an image, the RoI point cloud is derived through the estimated depth map. 
	Then the cropped RoI 3D space can be divided into a voxel grid of shape $n_x\times n_y\times n_z$, along the $X,Y,Z$ axis, respectively, where voxel sizes are determined according to:
\begin{equation}
S_{i\in{(x,y,z)}}= \frac{\max(P_i)-\min(P_i)}{n_i},
\end{equation}
 in which $i$ refers to the axis, $S$ is the voxel size, and $P$ is the 3D coordinates of the RoI point cloud. 
	We can see that $(\max(P_i)-\min(P_i))$ represents 3D spatial offsets along each axis in the RoI point cloud. 
	It indicates that voxel sizes are determined by the local spatial distribution and can adapt online to different objects to encode information better. 
	In light of this, we term it as object-aware adaptive size (O-A). 

	Nevertheless, the object-aware adaptive size is hard to explore the in-depth spatial distribution inside objects. 
	In the noisy RoI point cloud, points can be dense somewhere while sparse elsewhere, resulting in an unbalanced feature distribution in the voxel grid. 
	To effectively handle this, based on object-aware adaptive size, another type of adaptive voxel size is introduced, where the size of each voxel in the voxel grid can be different to fit the data distribution.
\begin{equation}
\small
\left\{
\begin{aligned}
MX_{k\in{\{0,1 \dots, n_x \}}}= {\rm sort}(P_x)\left[\lfloor{k\frac{N}{n_x}}\rfloor\right]\\
MY_{k\in{\{0,1 \dots, n_y \}}}={\rm  sort}(P_y)\left[\lfloor{k\frac{N}{n_y}}\rfloor\right]\\
MZ_{k\in{\{0,1 \dots, n_z \}}}= {\rm sort}(P_z)\left[\lfloor{k\frac{N}{n_z}}\rfloor\right]
\end{aligned}   
\right.
\label{equ:point-level}
\end{equation}

	As shown in Eq. \ref{equ:point-level} and Fig. \ref{fig:interval}, ${\rm sort}$ is the sort operation among 3D coordinates (\eg, ${\rm sort}(P_x)$ refers to the sorted $x$ coordinates of the RoI point cloud), $N$ is the number of points, and $\left[\cdot\right]$ denotes gathering the corresponding index.
	$MX$ represents the $x$ coordinates with given indexes in the sorted RoI point cloud. $MY$ and $MZ$ follow this line.
	More specifically, $MX,MY,MZ$ refer to retrieve a series of coordinates that locate on specified indexes, dividing sorted coordinates into $n_x,n_y,n_z$ parts.
	Therefore, the size of each voxel in the voxel grid depends on the spatial offset between head elements of two adjacent sorted parts (\eg, the $k$th voxel size along $X$ axis is $MX_{k+1} - MX_{k}$), making it varies a lot inside the unevenly distributed RoI point cloud. 
	The region with dense points more likely belongs to an object, and it will be partitioned with small voxel sizes, enabling more information to be maintained. 
	By contrast, the region with sparse points that are more likely outliers will be partitioned with large voxel sizes, generating sparse voxels, allowing more irrelevant interferences to be eliminated.
	We call this type of voxel size the point-aware adaptive size (P-A). 
	
	Aiming to obtain the 3D object center, as shown in Fig. \ref{fig:net2}, the location network acquires the adaptive voxels and outputs the location heat map with the same shape of input, where the 3D object center is the location with max probability.
	The heat map groundtruth is generated as described in \cite{CenterNet}, and smooth L1 loss is adopted during the training.
	Also, to avoid the object-aware adaptive size is heavily affected by outlier points, we remove these points whose depths are much larger than the average depth \cite{AM3D}.

\begin{figure}
\begin{minipage}[b]{0.48\textwidth} 
\centering 
			\subfigure[]{
				\label{fig_1}
				\centering
				\includegraphics[width=0.43\linewidth]{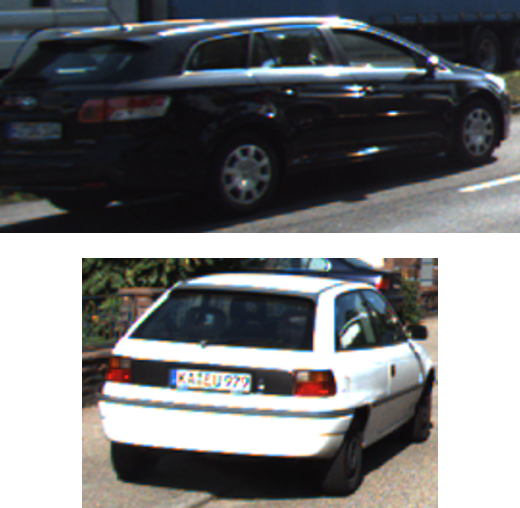}      
			}
			\subfigure[]{
			\label{fig_2}
				\centering
				\includegraphics[width=0.2\linewidth]{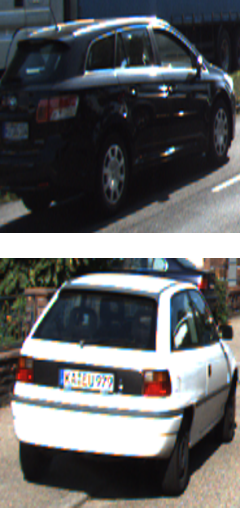}  
			}
			\subfigure[]{
				\label{fig_3}
				\centering
				\includegraphics[width=0.2\linewidth]{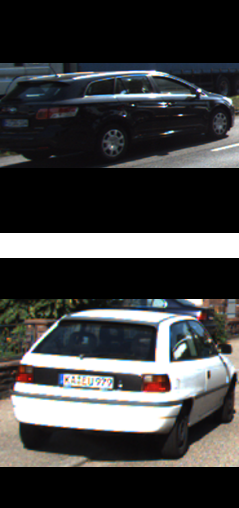}  
			}
		\caption{Illustration of different resizing schemes. (a): the original cropped {\bf \underline o}bject  {\bf \underline i}mage  {\bf \underline p}atch (OIP), (b): OIP after naive resizing, (c): OIP after Shape-Retaining(S-R).}
		\label{fig:shape}
\end{minipage}
\end{figure}

\subsection{Input for Orientation Estimation} \label{sec:orient}
	The orientation in 3D object detection contains the global orientation and the observation angle (local orientation), where the former is defined under the world space and the observation angle is defined under the camera coordinates.
	The global orientation is the goal and can be derived from the observation angle and the camera viewing angle. 
	
	Most state-of-the-art methods \cite{M3D,MonoGRNet,AM3D,D4LCN} directly infer the observation angle via deep features extracted from an entire image or a noisy point cloud. 
	However, the observation angle can be better predicted via the local RoI information as it only hinges on the object image patch. 
	Information from context does not offer much help because semantics from the region outside the object may bring irrelevant or even misleading information for local semantics.
	Also, the noisy point cloud can not express the object orientation precisely due to the noisy nature.
	Thus we argue that the cropped image patch alone is preferable as the input for orientation estimation.
		
	Interestingly, some prior works such as \cite{Deep3DBBox} also utilize the cropped image patch to achieve the orientation. 
	However, they are unaware of the underlying mechanism and improperly use a naive resizing scheme to resize object image patches to fit the network input.
	As shown in Fig. \ref{fig:shape}, naive resizing may destroy the visual semantics and thus tends to make the observation angles ambiguous.
	To this aim, we introduce another simple yet angle-preserving resizing scheme, called Shape-Retaining (S-R), i.e., resizing to match larger edge while keeping aspect ratio and then padding zeros symmetrically to get the desired shape. 
	In this way, information for the observation angle is maintained without damaging, thus better prediction accuracy can be achieved, as shown in Tab. \ref{tab:AOS_pes} and Tab. \ref{tab:AOS}.
	We obviously outperforms other monocular methods and even gives comparable results to the LiDAR-based method  \cite{FPointnet}  on the pedestrian category.

	\begin{figure*}[hbtp]
		\centering
				\includegraphics[width=1\linewidth]{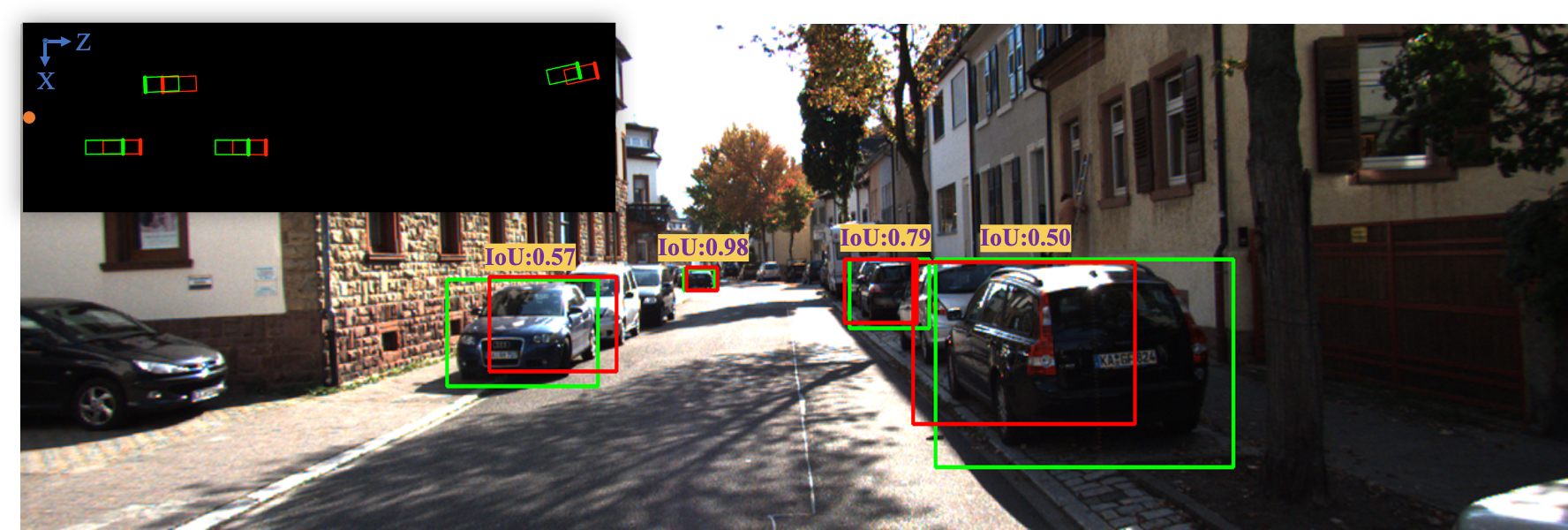}
		 \caption{IoU of 3D boxes on image plane at different distances. Top left refers to the BEV map and the boxes in the RGB image are the projections. With the growth of distance, the IoU of 3D boxes on the image plane becomes larger, although their offset keeps unchanged (2 meters here). }
		\label{fig:conf}
	\end{figure*}

\subsection{Decomposed 3D Detection Confidence} \label{sec:conf}
	In 2D detection, the confidence usually denotes the classification score to rate a box, which can be naturally learned in 2D detection. 
	An object assigned with a high confidence score means that it can be easily discriminated from the 2D image, but does not necessarily mean that it can be easily located accurately in 3D space. 
	For example, truncated/occluded cars, which can be easily detected in 2D, are considered as hard cases in 3D. 
	Therefore, directly adopting 2D confidence as 3D confidence, like many prior works \cite{PseudoLidar,Mono_PseudoLidar,AM3D}, are not suggested.
	Also, the 3D detection confidence can be learned during training by designing a specific loss using the 3D IoU \cite{FQNet}, but this manner is indirect and ignores the geometry relationship of 3D boxes and associated 2D boxes.
	To make estimation of 3D detection confidence more accurately, we decompose it as follows:
	\begin{equation}
	Conf_{3D}={Conf_{2D}}\times{Conf_{2D\rightarrow3D}}
\label{conf1}
\end{equation}
	The former is just the 2D detection confidence, and the latter measures the lifting hardness of the object from 2D to 3D, which considers how difficulty in locating a known 2D object in 3D space.
	Specifically, we achieve the lifting hardness by using the 2D-3D box IoU as the fact that the precise 3D box projections should fit its corresponding 2D box tightly, which is a much simpler yet more effective method.
	In particular, with the growth of the object distance, we find that the 2D-3D box IoU becomes larger for the object with the same location error, which means that the distance should also be taken into consideration. 
Fig. \ref{fig:conf} shows some examples. 
	Based on the above analysis, the proposed 3D detection confidence is calculated as follows:
\begin{equation}
Conf_{2D\rightarrow3D}=\frac{{\rm IoU}(box_{3D-proj}, box_{2D})}{e^{dis/\lambda}}
\label{conf2}
\end{equation}
	in which $dis$ refers to the distance from the predicted object to the camera optical center, {\it i.e.}, $\sqrt{x^2+y^2+z^2}$, $Conf_{2D\rightarrow3D}$ denotes the lifting hardness from 2D to 3D,  $\lambda$ is used to scale the depth, and $box_{3D-proj}$ refers to the 2D bounding box on predicted 3D box projections.
	When the distance becomes larger, the estimation error of location is increasing exponentially (see the supplementary material) as the information for location is less rich. 
	Therefore, we adopt the exponential transformation for the distance in Eq. \ref{conf2} to account for location error implicitly.
	In this way, we balance influences brought  from the 2D-3D box IoU and the location.
	It is worthy to note that our method can be plugged into any other 2D-detector-based monocular method without cost to boost the performance.

	\begin{table*}
		\begin{center}
			\begin{tabular}{l|ccc|ccc}
				\toprule  
				\multirow{2}{*}{Approach} &  \multicolumn{3}{c|}{AP$_{BEV}$/AP$_{3D}$ (IoU=0.5)$|\scriptstyle R_{11}$}  &  \multicolumn{3}{c}{AP$_{BEV}$ /AP$_{3D}$ (IoU=0.7)$|\scriptstyle R_{11}$} \\ 
				~ &  Easy & Moderate & Hard & Easy & Moderate & Hard\\ 
				\midrule         
				OFTNet \cite{OFTNet}   &  -     &  -     & -  &  11.06/4.07   &   8.79/3.27  & 8.91/3.29   \\ 
				RoI-10D \cite{ROI10D}  &  -     &  -     & -  & 14.50/10.25    &   9.91/6.39   & 8.73/6.18 \\ 
				MonoDIS \cite{AP40}  & -     &  -     & -   &  24.26/18.05   & 18.43/14.98     & 16.95/13.42 \\ 
				MonoGRNet \cite{MonoGRNet}  &  -/50.51  & -/36.97     & -/30.82  & -/13.88  & -/10.19 & -/7.62\\
				Deep3DBBox \cite{Deep3DBBox}  & 30.02/27.04   &  23.77/20.55   & 18.83/15.88  &  9.99/5.85  & 7.71/4.10  & 5.30/3.84 \\
				Mono3D \cite{Mono3D}  &  30.50/25.19   & 22.39/18.20     & 19.16/15.52 &  5.22/2.53   &  5.19/2.31    &  4.13/2.31 \\ 
				FQNet \cite{FQNet}  &  32.57/28.16  &  24.60/21.02  & 21.25/19.91  & 9.50/5.98    & 8.02/5.50    &  7.71/4.75 \\ 
				MonoPSR \cite{MonoPSR} &56.97/49.65 &43.39/41.71 & 36.00/29.95 & 20.63/12.75 & 18.67/11.48 & 14.45/8.59\\
				M3D-RPN \cite{M3D}  & 55.37/48.96  & 42.49/39.57  & 35.29/33.01 &  25.94/20.27  & 21.18/17.06 &17.90/15.21 \\ 
				Pseudo-LiDAR \cite{PseudoLidar}  &  \underline{63.42}/56.54  & 40.86/37.53   & 37.69/32.24 & 31.88/{\underline{24.12}}   &  20.84/15.74  &18.92/14.96\\
				D4LCN \cite{D4LCN}   &  54.35/51.30 & 40.33/35.10  & 33.96/32.46   & 26.00/19.38  &  20.73/16.00   & 17.46/12.94 \\
				RTM3D \cite{RTM3D} & 57.47/54.36 & \underline{44.16}/\underline{41.90} & {\underline{42.31}}/\underline{35.84} & {25.56/20.77} & \underline{22.12}/{16.86} & \underline{20.91}/{\bf 16.63} \\
				PatchNet \cite{PatchNet} & {\bf64.87}/{\bf58.63} & {40.94/38.20} & {38.16//32.50} & \underline{32.30}/{\bf {25.76}} & {21.25}/\underline{17.72} & {19.04/15.62} \\
				\midrule 
				{\bf Ours} & 61.78/{\underline{58.07}}   & \bf{45.65/42.61}   & \bf{42.32}/\bf{36.62}  & {\bf 33.24}/23.65   &  {\bf{24.25}}/\bf{17.75}   &  {\bf21.67}/\underline{15.93} \\

				\bottomrule 
			\end{tabular}
		\end{center}
		\caption{Comparison of our method to other monocular 3D object detection methods for car category on KITTI validation set. For a fair comparison, depth map based methods \cite{PseudoLidar,D4LCN,PatchNet} are retrained using our proposed depth dataset split. To include more methods in comparison, all the methods are evaluated with metric $AP|_{R_{11}}$.}
		\label{tab:quantitative_results}
	\end{table*}

\section{Experiments}\label{sec:Experiments}

\subsection{Implementation Details}
	We implement our framework using Pytorch \cite{Pytorch} and on a Nvidia 1080Ti GPU, applying Adam optimizer \cite{Adam}, in which the learning rate is set to $10^{-4}$. 
	The encoder-decoder network architecture is employed for the location sub-network and is trained with a total of 50 epochs (see the supplementary material for more details). 
	The location results are refined by the 2D-3D box consistency.
	For the orientation and dimension sub-network, we use the protocol introduced in \cite{Deep3DBBox}. 
	Depth estimator and 2D detector are brought from \cite{BTS} and \cite{FPointnet}, using our remedied depth dataset split to train the depth estimator. 
	The transformed point cloud is rotated as described in \cite{FPointnet}, and the shape of the voxel grid ($n_x,n_y,n_z$) is set to $(32, 16, 64)$.
	$\lambda$ in Eq. \ref{conf2} is 80 in default since 80 meters is the usual max depth in KITTI. 
	We provide extra experimental results in the supplementary material due to the space limitation.

  \begin{table}
        \centering
\footnotesize
			\begin{tabular}{p{0.15cm}<{\centering}p{0.15cm}<{\centering}p{0.8cm}<{\centering}p{0.3cm}<{\centering}|ccr}
				\toprule  

				 \multirow{2}{*}{O-A}& \multirow{2}{*}{S-R}& \multirow{2}{*}{3D-Conf} & \multirow{2}{*}{P-A}  &  \multicolumn{3}{c}{AP$_{BEV}$/AP$_{3D}$ (IoU=0.7)$|\scriptstyle R_{40}$} \\ 
				~ & ~ & ~ & ~ & Easy & Moderate & Hard\\   
				\midrule     
					$\surd$ & ~  &~    &~  &22.12/12.37   & 11.89/6.08  & 8.77/4.41   \\ 
					$\surd$ & $\surd$  &~   &~  &24.45/14.60   & 12.78/7.20  & 9.37/5.31   \\ 
				$\surd$  &  $\surd$  &  $\surd$  &~  &27.93/17.36   & 16.52/10.04  & 12.04/7.25  \\   
				~ &  $\surd$   &$\surd$    &$\surd$  &28.13/17.57   & 19.90/\bf{12.10}  & 16.86/{\bf10.36}  \\   
				 $\surd$  & $\surd$  & $\surd$   & $\surd$  &\bf{29.57/18.08}    &  {\bf20.01}/12.09 & {\bf16.89}/10.15  \\   
				 
				\bottomrule 
			\end{tabular}
						      \caption{Ablation study.}
			\label{tab:ablation}
      \end{table}

 \begin{table}
  \footnotesize
      \centering
			\begin{tabular}{l|ccc}
				\toprule   
				\multirow{2}{*}{\tabincell{l}{3D Confidence}} &  \multicolumn{3}{c}{AP$_{BEV}/$AP$_{3D}$ (IoU=0.7)$|\scriptstyle R_{40}$} \\ 
				~ & Easy & Moderate & Hard\\ 
				\midrule         
				Learning-based  & 26.85/15.39 & 16.41/9.33 & 14.70/9.02 \\
				\midrule 
				{\bf Projection-based}  &\bf{29.57/18.08}  &  \bf{20.01/12.09} & {\bf16.89}/{\bf 10.15}\\
				\bottomrule 
			\end{tabular}
			\caption{Comparisons on different 3D confidence schemes.}
			\label{tab:conf3d}
  \end{table}

\subsection{Dataset and Metrics}
 \noindent
{\bf{KITTI Dataset:}}
	The KITTI dataset is widely used for monocular 3D object detection.
	KITTI \cite{KITTI2012} provides 7,481 images for training and 7,518 images for testing. 
	The test set is confidential and can only be tested on the KITTI website, while the training set is publicly available. 
	To keep consistent with previous works, we adopt the dataset split described in \cite{3DOP} for comprehensive experiments.
	The available data is divided into a training set (3,712 images) and a validation set (3,769 images).   

 ~\\
 \noindent
{\bf{Depth Dataset Split for 3D Object Detection:}}{\label{sec:split}}
	Existing depth estimation methods usually are trained upon the split proposed in \cite{Eigen}, in which 32 scenes among all KITTI scenarios are used for training. 
	To the best of our knowledge, most depth-based monocular 3D object detection methods estimate depth directly from the existing pre-trained models. 
	However, the training set for depth estimation overlaps with the validation set for 3D object detection, resulting in overrating these methods. 
	To eliminate this problem, we introduce a new depth dataset split for 3D object detection. 
	Specifically, we exclude scenes that emerge in the KITTI 3D object validation set. 
	The remaining scenes consist of the new training set, and others are divided into the validation set.
	This depth dataset split contains data from non-overlapping sequences and fixes the data leakage issue. 
	Thus we re-train and re-evaluate depth-based methods \cite{PseudoLidar,D4LCN,PatchNet} using the official publicly available code.

 ~\\
 \noindent
{\bf{Evaluation Metrics:}}
	We conduct experiments on the KITTI validation set and official test set under two core tasks: bird's eye view (BEV) and 3D object detection in three difficulties. 
	Difficulties of objects are subdivided into easy, moderate, and hard according to the occlusion level, truncation, and bounding box height.
	Many previous works evaluate their results with $AP_{11}$, while $AP_{40}$ proposed in \cite{AP40} is suggested to be adopted by the official KITTI benchmark recently. 
	Therefore we provide performances under both $AP_{11}$ and $AP_{40}$ metrics on the car category to comprehensively evaluate the proposed method.  

\subsection{Quantitative Results}
	We compare our method with current state-of-the-art monocular methods in the KITTI validation set. 
	Note that depth map based methods are trained upon the new depth dataset split. 
	As shown in Tab. \ref{tab:quantitative_results}, our method achieves new state-of-the-art results. 
	For the BEV and 3D tasks, we gain significant improvements compared to other state-of-the-art results. 
	Also, we provide comparisons on KITTI testing set as shown in Tab. \ref{tab:test}, in which our method still achieves state-of-the-art.
	It is worthy to note that results of our method should not be compared directly with other depth map based methods which utilize a different depth estimator trained with the un-remedied depth dataset split.

  \begin{table}
      \centering
      \footnotesize
			\begin{tabular}{l|ccc}
				\toprule   
				Approach &Easy & Moderate & Hard \\ 
				\midrule         
				AM3D \cite{AM3D}* & 25.03/16.50 & 17.32/10.74 & 14.91/9.52\\
				D4LCN \cite{D4LCN}*  & 22.51/16.65 & 16.02/11.72 & 12.55/9.51 \\
				PatchNet \cite{PatchNet}* & 22.97/15.68 & 16.86/11.12 & 14.97/10.17 \\
				\midrule 
				ROI-10D \cite{ROI10D} & 9.78/4.32  & 4.91/2.02  & 3.74/1.46 \\ 
				MonoGRNet \cite{MonoGRNet} & 18.19/5.74  & 11.17/9.61  & 8.73/4.25 \\
				MonoPSR \cite{MonoPSR} & 18.33/10.76 & 12.58/7.25 & 9.91/5.85 \\
				M3D-RPN \cite{M3D}  &{21.02/14.76}  & 13.67/9.71  & 10.23/7.42   \\
				MonoPair \cite{MonoPair} & 19.28/13.04 & {14.83}/9.99 & {12.89}/8.65 \\
				RTM3D \cite{RTM3D} & 19.17/14.41  & 14.20/{10.34}  &  11.99/{\bf 8.77}    \\ 
				\midrule
				{\bf Ours}  & {\bf 27.87/17.48 }   &  {\bf 17.13/10.44}    &  {\bf 13.53}/{{7.87}}\\
				\bottomrule 
			\end{tabular}
			\caption{
			Comparisons on KITTI testing set. Note that $*$ denotes the method use a different depth estimator with different depth dataset split compared to ours.
			}
			\label{tab:test}
  \end{table}

\subsection{Detailed Analysis}
 \noindent
{\bf{Ablation Study:}}
	We provide extensive experiments to study the impact of each critical component in our framework. 
	As shown in Tab. \ref{tab:ablation}, modules: object-aware adaptive voxel size (O-A), point-aware adaptive voxel size (P-A),  Shape-Retaining (S-R), decomposed 3D confidence (3D-Conf) are gradually added to the framework, producing growing improvements. 
	We also compare different decomposed 3D confidence schemes mentioned in Sec. \ref{sec:conf}, i.e., the learning-based method using 3D IoU \cite{FQNet} and our 2D-3D projection-based method.
	We can the that the simple yet effective projection-based method performs much better.

    	\begin{table*}
	\scriptsize
		\begin{center}
			\begin{tabular}{l|c|p{1.05cm}<{\centering}p{1.05cm}<{\centering}p{1.25cm}<{\centering}|p{1.0cm}<{\centering}p{1.0cm}<{\centering}p{1.2cm}<{\centering}|p{1.0cm}<{\centering}p{1.0cm}<{\centering}p{1.0cm}<{\centering}} 
				\toprule  
				\multirow{2}{*}{Approaches/Input}  & \multirow{2}{*}{Dataset} & \multicolumn{3}{c|}{AOS$|\scriptstyle R_{40}$~~~} & \multicolumn{3}{c}{AP$_{BEV}$$|\scriptstyle R_{40}$~~~}  &  \multicolumn{3}{c}{AP$_{3D}$$|\scriptstyle R_{40}$} \\ 
				~ &  ~ & Easy & Mod & Hard & Easy & Mod & Hard & Easy & Mod & Hard\\ 
				\midrule
				M3D-RPN \cite{M3D} /Entire image     &\multirow{2}{*}{\tabincell{c}{ Valida- \\ tion set}}      & 44.86  &  37.57  & 31.72 & 5.18 &4.48 &3.61  & 4.75 &3.55&2.79   \\ 
				{\bf M3D-RPN/Image patch(Ours)} & ~ & \bf  \textcolor{red}{63.01(+18.15)}  &   \bf  \textcolor{red}{53.11(+16.54)}  &  \bf  \textcolor{red}{44.69(+12.97)}  & \bf   \textcolor{red}{6.02(+0.84)}   & \bf  \textcolor{red}{4.79(+0.31)}   & \bf  \textcolor{red}{4.03(+0.42)}  & \bf   \textcolor{red}{5.24(+0.49)}   & \bf  \textcolor{red}{4.33(+0.78)}  & \bf  \textcolor{red}{3.63(+0.84)}  \\
				
				\midrule
				M3D-RPN \cite{M3D} /Entire image     &\multirow{2}{*}{\tabincell{c}{ Test \\ set}}      & 44.33  &  31.88  & 28.55 &5.65 &4.05 &	3.29  & 4.92 &3.48 &2.94   \\ 
				{\bf M3D-RPN/Image patch(Ours)} & ~ & \bf  \textcolor{red}{51.90(+7.57)}  &   \bf  \textcolor{red}{37.78(+5.90)}  &  \bf  \textcolor{red}{33.95(+5.40)}  & \bf   \textcolor{red}{6.53(+0.98)}   & \bf  \textcolor{red}{4.46(+0.39)}   & \bf  \textcolor{red}{4.10(+0.91)}  & \bf   \textcolor{red}{5.70(+0.78)}   & \bf  \textcolor{red}{4.11(+0.63)}  & \bf  \textcolor{red}{3.37(+0.43)}  \\
				\midrule         
				{\bf F-PointNet \cite{FPointnet} /LiDAR }  &  \multirow{2}{*}{\tabincell{c}{ Valida- \\ tion set}}    & \bf  \textcolor{red}{81.18(+1.05)} &\bf  \textcolor{red}{72.06+(0.01)} & 66.00  & \bf  \textcolor{red}{72.15+(1.79)}  & \bf  \textcolor{red}{ 62.16+(1.54)} & \bf  \textcolor{red}{54.13+(0.80)}& \bf  \textcolor{red}{64.99+(1.43)} &\bf  \textcolor{red}{55.12+(1.65)} &	\bf  \textcolor{red}{47.26+(0.87)} \\
				F-PointNet/Image patch(Ours)& ~ & 80.13 & 72.05 & \bf  \textcolor{red}{66.05+(0.05)} & 70.36 & 60.52 & 53.33 & 63.56 & 53.47 & 46.39\\
				
				\bottomrule 
			\end{tabular}
		\end{center}
		\caption{Comparison of different inputs for estimating orientation on the pedestrian category on KITTI benchmark. Our method gives much better results and even be comparable with the LiDAR-based method \cite{FPointnet}.}
		\label{tab:AOS_pes}
	\end{table*}

	  \begin{table}[hbtp]
  \footnotesize
      \centering
			\begin{tabular}{l|ccc}
				\toprule  
			\multirow{2}{*}{Methods} &  \multicolumn{3}{c}{AP$_{BEV}/$AP$_{3D}$ (IoU=0.7)$|\scriptstyle R_{40}$} \\ 
				~ & Easy & Moderate & Hard\\ 
				\midrule         
				P-LiDAR \cite{PseudoLidar} & 27.29/18.43    & 15.10/9.74  & 12.61/7.81 \\ 
				P-LiDAR+3D-Conf & \bf 31.77/21.48    & \bf 21.12/14.36  & \bf 17.73/11.67\\ 
				\midrule
				PatchNet \cite{PatchNet}  & 27.97/20.75    &  15.67/11.26 &  12.92/9.07 \\
				PatchNet+3D-Conf  & \bf 31.52/23.57    & \bf 20.94/15.85 & \bf 17.40/12.78 \\
				\bottomrule 
			\end{tabular}
		      \caption{Extensibility of our decomposed 3D confidence scheme. "P-LiDAR" in the table denotes Pseudo-LiDAR \cite{PseudoLidar}. The proposed 3D confidence scheme brings significant performance improvements.}
			\label{tab:3D-conf}
      \end{table}

~\\
 \noindent
{\bf{Extensibility of Decomposed 3D Confidence Scheme:}}
 	As mentioned in Sec. \ref{sec:conf}, our decomposed 3D confidence scheme only requires the 2D detection confidence and the predicted 3D detection results, meaning that it can be easily plugged into any monocular 3D object detection method based on a 2D detector. 
	As shown in Tab. \ref{tab:3D-conf}, by employing our decomposed 3D confidence scheme, Pseudo-LiDAR \cite{PseudoLidar} gains {\bf 39.87\%/47.4\%} relative improvements, and PatchNet \cite{PatchNet} gains {\bf 33.63\%/40.76\%} relative improvements in {\it moderate} setting. 
	This significant boosting demonstrates the effectiveness of our decomposed 3D confidence scheme.

  \begin{table}
        \centering
    \begin{minipage}{0.45\textwidth}
      \centering
 	 \footnotesize
			\begin{tabular}{lc|ccc}
				\toprule  
			Method &Metric &  Easy & Moderate & Hard \\
				\midrule         
				\multirow{3}{*}{\tabincell{l}{M3D-RPN \cite{M3D}}} &AOS(Org.)  & 89.37  &  81.58  & 65.74   \\ 	
				~ &  AOS(Ours)  & \bf{89.42}    &  \bf{82.53}  & \bf{66.64} \\
				~ &2D Det. & 90.02    &  83.14   & 67.37\\ 		
				\midrule         
				\multirow{3}{*}{D4LCN \cite{D4LCN}} & AOS(Org.)  & 91.75    &  82.97  & 66.45\\ 
				~ &AOS(Ours) & \bf{92.12}    &  \bf{83.27}  & \bf{66.81}  \\
				~ & 2D Det. & 92.76    &  84.40   & 67.86 \\ 		
				\midrule         
				\multirow{3}{*}{\tabincell{l}{Pseudo-\\LiDAR \cite{PseudoLidar}}} & AOS(Org.)  & 92.85    &  82.29  & 78.51\\ 
				~ &AOS(Ours) & \bf{96.15}    &  \bf{89.84}  & \bf{85.56}  \\
				~ & 2D Det. & 96.48    &  90.30   & 87.62 \\ 	
				\bottomrule 
			\end{tabular}
			\caption{Comparison of different way to estimate orientation on car category on the KITTI validation set. "2D Det." in the table denotes the AP of 2D detection, which is the up-bound for AOS.}
			\label{tab:AOS}
    \end{minipage}
  \end{table}

 ~\\
 \noindent
{\bf{Different Inputs for Orientation Estimation:}}
	In Sec. \ref{sec:orient}, we argue that the object image patch alone provide enough information for orientations of objects in the image, which is proved here via quantitative experiments.
	Prior state-of-the-art (SOTA) methods, M3D-RPN \cite{M3D}, D4LCN \cite{D4LCN}, Pseudo-LiDAR \cite{PseudoLidar} are chosen as the baseline. 
	They predict orientations via extracted features from the entire image or the point cloud. 
	By contrast, we only use the same 2D boxes produced by them and apply our method.
	We show the results of Average Orientation Similarity (AOS) on the car category in Tab. \ref{tab:AOS}, where AOS(Org.) denotes performances produced by the original method and 2D Det. denotes the AP of 2D detection, which is the up-bound for AOS.
	Our method performs much better than the original method with a significant gap, demonstrating our perspective. 
	Besides, we conduct experiments on the pedestrian category as shown in Tab. \ref{tab:AOS_pes}. 
	By replacing only our orientations to M3D-RPN \cite{M3D}, dramatic improvements are obtained both in the validation set and test set. 
	Even if compared to the LiDAR-based method \cite{FPointnet}, it achieves comparable performances.
	These huge gains can be attributed to that our method maintains the information as much as possible and makes the model focus on the object image patch. 
	Especially for the pedestrian category, which has a high aspect ratio, previous methods bring in interferences of context, thus perform much worse.

\subsection{Limitations}
	Our current framework still has some limitations. 
	First, compared to other end-to-end algorithms such as \cite{M3D}, our training process is of complication. 
	All sub-modules can be integrated and trained with an end-to-end learning approach.
	Second, we have not taken some more useful information into consideration, such as semantics. 
	Such aspects will be explored in our future work.

\section{Conclusions}
	In this paper, we propose a novel data representation to encode the depth map and the RGB image in which voxel sizes are adaptively determined by the spatial distribution of the transformed point cloud. 
	This representation is used to predict objects' locations. 
	Furthermore, we propose that the object image patch alone with a proper resizing scheme is a better input than other methods that utilize the entire image or the noisy point cloud. Experiments demonstrate our perspective. 
	Besides, we introduce a novel decomposed 3D confidence approach, constructing the connection between 2D detection and 3D detection.
	Finally, we reveal a data-leakage problem and propose a new dataset split to fix it. 
	Extensive experiments demonstrate that our method is promising and outperforms other state-of-the-art methods.

{\small
\bibliographystyle{ieee_fullname}
\bibliography{egbib}

\begin{thebibliography}{10}\itemsep=-1pt

\bibitem{M3D}
Garrick Brazil and Xiaoming Liu.
\newblock M3d-rpn: Monocular 3d region proposal network for object detection.
\newblock In {\em Proceedings of the IEEE International Conference on Computer
  Vision}, pages 9287--9296, 2019.

\bibitem{Kinematic3D}
Garrick Brazil, Gerard Pons-Moll, Xiaoming Liu, and Bernt Schiele.
\newblock Kinematic 3d object detection in monocular video.
\newblock {\em arXiv preprint arXiv:2007.09548}, 2020.

\bibitem{Mono3D}
Xiaozhi Chen, Kaustav Kundu, Ziyu Zhang, Huimin Ma, Sanja Fidler, and Raquel
  Urtasun.
\newblock Monocular 3d object detection for autonomous driving.
\newblock In {\em Proceedings of the IEEE Conference on Computer Vision and
  Pattern Recognition}, pages 2147--2156, 2016.

\bibitem{3DOP}
Xiaozhi Chen, Kaustav Kundu, Yukun Zhu, Huimin Ma, Sanja Fidler, and Raquel
  Urtasun.
\newblock 3d object proposals using stereo imagery for accurate object class
  detection.
\newblock {\em IEEE transactions on pattern analysis and machine intelligence},
  40(5):1259--1272, 2017.

\bibitem{DSGN}
Yilun Chen, Shu Liu, Xiaoyong Shen, and Jiaya Jia.
\newblock Dsgn: Deep stereo geometry network for 3d object detection.
\newblock In {\em Proceedings of the IEEE/CVF Conference on Computer Vision and
  Pattern Recognition}, pages 12536--12545, 2020.

\bibitem{MonoPair}
Yongjian Chen, Lei Tai, Kai Sun, and Mingyang Li.
\newblock Monopair: Monocular 3d object detection using pairwise spatial
  relationships.
\newblock In {\em Proceedings of the IEEE/CVF Conference on Computer Vision and
  Pattern Recognition}, pages 12093--12102, 2020.

\bibitem{3D_UNet}
{\"O}zg{\"u}n {\c{C}}i{\c{c}}ek, Ahmed Abdulkadir, Soeren~S Lienkamp, Thomas
  Brox, and Olaf Ronneberger.
\newblock 3d u-net: learning dense volumetric segmentation from sparse
  annotation.
\newblock In {\em International conference on medical image computing and
  computer-assisted intervention}, pages 424--432. Springer, 2016.

\bibitem{D4LCN}
Mingyu Ding, Yuqi Huo, Hongwei Yi, Zhe Wang, Jianping Shi, Zhiwu Lu, and Ping
  Luo.
\newblock Learning depth-guided convolutions for monocular 3d object detection.
\newblock In {\em Proceedings of the IEEE/CVF Conference on Computer Vision and
  Pattern Recognition}, pages 11672--11681, 2020.

\bibitem{Eigen}
David Eigen, Christian Puhrsch, and Rob Fergus.
\newblock Depth map prediction from a single image using a multi-scale deep
  network.
\newblock In {\em Advances in neural information processing systems}, pages
  2366--2374, 2014.

\bibitem{KITTI2012}
Andreas Geiger, Philip Lenz, and Raquel Urtasun.
\newblock Are we ready for autonomous driving? the kitti vision benchmark
  suite.
\newblock In {\em 2012 IEEE Conference on Computer Vision and Pattern
  Recognition}, pages 3354--3361. IEEE, 2012.

\bibitem{SA-SSD}
Chenhang He, Hui Zeng, Jianqiang Huang, Xian-Sheng Hua, and Lei Zhang.
\newblock Structure aware single-stage 3d object detection from point cloud.
\newblock In {\em Proceedings of the IEEE/CVF Conference on Computer Vision and
  Pattern Recognition}, pages 11873--11882, 2020.

\bibitem{Epnet}
Tengteng Huang, Zhe Liu, Xiwu Chen, and Xiang Bai.
\newblock Epnet: Enhancing point features with image semantics for 3d object
  detection.
\newblock {\em arXiv preprint arXiv:2007.08856}, 2020.

\bibitem{Adam}
Diederik~P Kingma and Jimmy Ba.
\newblock Adam: A method for stochastic optimization.
\newblock {\em arXiv preprint arXiv:1412.6980}, 2014.

\bibitem{MonoPSR}
Jason Ku, Alex~D Pon, and Steven~L Waslander.
\newblock Monocular 3d object detection leveraging accurate proposals and shape
  reconstruction.
\newblock In {\em Proceedings of the IEEE Conference on Computer Vision and
  Pattern Recognition}, pages 11867--11876, 2019.

\bibitem{PP}
Alex~H Lang, Sourabh Vora, Holger Caesar, Lubing Zhou, Jiong Yang, and Oscar
  Beijbom.
\newblock Pointpillars: Fast encoders for object detection from point clouds.
\newblock In {\em Proceedings of the IEEE Conference on Computer Vision and
  Pattern Recognition}, pages 12697--12705, 2019.

\bibitem{BTS}
Jin~Han Lee, Myung-Kyu Han, Dong~Wook Ko, and Il~Hong Suh.
\newblock From big to small: Multi-scale local planar guidance for monocular
  depth estimation.
\newblock {\em arXiv preprint arXiv:1907.10326}, 2019.

\bibitem{StereoRCNN}
Peiliang Li, Xiaozhi Chen, and Shaojie Shen.
\newblock Stereo r-cnn based 3d object detection for autonomous driving.
\newblock In {\em Proceedings of the IEEE Conference on Computer Vision and
  Pattern Recognition}, pages 7644--7652, 2019.

\bibitem{RTM3D}
Peixuan Li, Huaici Zhao, Pengfei Liu, and Feidao Cao.
\newblock Rtm3d: Real-time monocular 3d detection from object keypoints for
  autonomous driving.
\newblock {\em arXiv preprint arXiv:2001.03343}, 2020.

\bibitem{FQNet}
Lijie Liu, Jiwen Lu, Chunjing Xu, Qi Tian, and Jie Zhou.
\newblock Deep fitting degree scoring network for monocular 3d object
  detection.
\newblock In {\em Proceedings of the IEEE Conference on Computer Vision and
  Pattern Recognition}, pages 1057--1066, 2019.

\bibitem{PVCNN}
Zhijian Liu, Haotian Tang, Yujun Lin, and Song Han.
\newblock Point-voxel cnn for efficient 3d deep learning.
\newblock In {\em Advances in Neural Information Processing Systems}, pages
  965--975, 2019.

\bibitem{FCN}
Jonathan Long, Evan Shelhamer, and Trevor Darrell.
\newblock Fully convolutional networks for semantic segmentation.
\newblock In {\em Proceedings of the IEEE conference on computer vision and
  pattern recognition}, pages 3431--3440, 2015.

\bibitem{PatchNet}
Xinzhu Ma, Shinan Liu, Zhiyi Xia, Hongwen Zhang, Xingyu Zeng, and Wanli Ouyang.
\newblock Rethinking pseudo-lidar representation.
\newblock {\em arXiv preprint arXiv:2008.04582}, 2020.

\bibitem{AM3D}
Xinzhu Ma, Zhihui Wang, Haojie Li, Pengbo Zhang, Wanli Ouyang, and Xin Fan.
\newblock Accurate monocular 3d object detection via color-embedded 3d
  reconstruction for autonomous driving.
\newblock In {\em Proceedings of the IEEE International Conference on Computer
  Vision}, pages 6851--6860, 2019.

\bibitem{ROI10D}
Fabian Manhardt, Wadim Kehl, and Adrien Gaidon.
\newblock Roi-10d: Monocular lifting of 2d detection to 6d pose and metric
  shape.
\newblock In {\em Proceedings of the IEEE Conference on Computer Vision and
  Pattern Recognition}, pages 2069--2078, 2019.

\bibitem{Deep3DBBox}
Arsalan Mousavian, Dragomir Anguelov, John Flynn, and Jana Kosecka.
\newblock 3d bounding box estimation using deep learning and geometry.
\newblock In {\em Proceedings of the IEEE Conference on Computer Vision and
  Pattern Recognition}, pages 7074--7082, 2017.

\bibitem{Pytorch}
Adam Paszke, Sam Gross, Francisco Massa, Adam Lerer, James Bradbury, Gregory
  Chanan, Trevor Killeen, Zeming Lin, Natalia Gimelshein, Luca Antiga, et~al.
\newblock Pytorch: An imperative style, high-performance deep learning library.
\newblock In {\em Advances in neural information processing systems}, pages
  8026--8037, 2019.

\bibitem{FPointnet}
Charles~R Qi, Wei Liu, Chenxia Wu, Hao Su, and Leonidas~J Guibas.
\newblock Frustum pointnets for 3d object detection from rgb-d data.
\newblock In {\em Proceedings of the IEEE conference on computer vision and
  pattern recognition}, pages 918--927, 2018.

\bibitem{MonoGRNet}
Zengyi Qin, Jinglu Wang, and Yan Lu.
\newblock Monogrnet: A geometric reasoning network for monocular 3d object
  localization.
\newblock In {\em Proceedings of the AAAI Conference on Artificial
  Intelligence}, volume~33, pages 8851--8858, 2019.

\bibitem{OFTNet}
Thomas Roddick, Alex Kendall, and Roberto Cipolla.
\newblock Orthographic feature transform for monocular 3d object detection.
\newblock {\em arXiv preprint arXiv:1811.08188}, 2018.

\bibitem{UNet}
Olaf Ronneberger, Philipp Fischer, and Thomas Brox.
\newblock U-net: Convolutional networks for biomedical image segmentation.
\newblock In {\em International Conference on Medical image computing and
  computer-assisted intervention}, pages 234--241. Springer, 2015.

\bibitem{PVRCNN}
Shaoshuai Shi, Chaoxu Guo, Li Jiang, Zhe Wang, Jianping Shi, Xiaogang Wang, and
  Hongsheng Li.
\newblock Pv-rcnn: Point-voxel feature set abstraction for 3d object detection.
\newblock In {\em Proceedings of the IEEE/CVF Conference on Computer Vision and
  Pattern Recognition}, pages 10529--10538, 2020.

\bibitem{PART}
Shaoshuai Shi, Zhe Wang, Jianping Shi, Xiaogang Wang, and Hongsheng Li.
\newblock From points to parts: 3d object detection from point cloud with
  part-aware and part-aggregation network.
\newblock {\em arXiv preprint arXiv:1907.03670}, 2019.

\bibitem{Point-gnn}
Weijing Shi and Raj Rajkumar.
\newblock Point-gnn: Graph neural network for 3d object detection in a point
  cloud.
\newblock In {\em Proceedings of the IEEE/CVF Conference on Computer Vision and
  Pattern Recognition}, pages 1711--1719, 2020.

\bibitem{AP40}
Andrea Simonelli, Samuel~Rota Bulo, Lorenzo Porzi, Manuel L{\'o}pez-Antequera,
  and Peter Kontschieder.
\newblock Disentangling monocular 3d object detection.
\newblock In {\em Proceedings of the IEEE International Conference on Computer
  Vision}, pages 1991--1999, 2019.

\bibitem{VoxelFPN}
Bei Wang, Jianping An, and Jiayan Cao.
\newblock Voxel-fpn: multi-scale voxel feature aggregation in 3d object
  detection from point clouds.
\newblock {\em arXiv preprint arXiv:1907.05286}, 2019.

\bibitem{PseudoLidar}
Yan Wang, Wei-Lun Chao, Divyansh Garg, Bharath Hariharan, Mark Campbell, and
  Kilian~Q Weinberger.
\newblock Pseudo-lidar from visual depth estimation: Bridging the gap in 3d
  object detection for autonomous driving.
\newblock In {\em Proceedings of the IEEE Conference on Computer Vision and
  Pattern Recognition}, pages 8445--8453, 2019.

\bibitem{Mono_PseudoLidar}
Xinshuo Weng and Kris Kitani.
\newblock Monocular 3d object detection with pseudo-lidar point cloud.
\newblock In {\em Proceedings of the IEEE International Conference on Computer
  Vision Workshops}, pages 0--0, 2019.

\bibitem{3dssd}
Zetong Yang, Yanan Sun, Shu Liu, and Jiaya Jia.
\newblock 3dssd: Point-based 3d single stage object detector.
\newblock In {\em Proceedings of the IEEE/CVF Conference on Computer Vision and
  Pattern Recognition}, pages 11040--11048, 2020.

\bibitem{HVNet}
Maosheng Ye, Shuangjie Xu, and Tongyi Cao.
\newblock Hvnet: Hybrid voxel network for lidar based 3d object detection.
\newblock In {\em Proceedings of the IEEE/CVF Conference on Computer Vision and
  Pattern Recognition}, pages 1631--1640, 2020.

\bibitem{3D-CVF}
Jin~Hyeok Yoo, Yeocheol Kim, Ji~Song Kim, and Jun~Won Choi.
\newblock 3d-cvf: Generating joint camera and lidar features using cross-view
  spatial feature fusion for 3d object detection.
\newblock {\em arXiv preprint arXiv:2004.12636}, 2020.

\bibitem{CenterNet}
Xingyi Zhou, Dequan Wang, and Philipp Kr{\"a}henb{\"u}hl.
\newblock Objects as points.
\newblock {\em arXiv preprint arXiv:1904.07850}, 2019.

\bibitem{VoxelNet}
Yin Zhou and Oncel Tuzel.
\newblock Voxelnet: End-to-end learning for point cloud based 3d object
  detection.
\newblock In {\em Proceedings of the IEEE Conference on Computer Vision and
  Pattern Recognition}, pages 4490--4499, 2018.

\end{thebibliography}
}

\end{document}